# WiFi-based Multi-task Sensing

Xie Zhang[1], Chengpei Tang[1], Yasong An[1], and Kang Yin[1]

[1] Sun Yat-sen University, Guangzhou 510006, China

**Abstract.** WiFi-based sensing has aroused immense attention over recent years. The rationale is that the signal fluctuations caused by humans carry the information of human behavior which can be extracted from the channel state information of WiFi. Still, the prior studies mainly focus on single-task sensing (STS), e.g., gesture recognition, indoor localization, user identification. Since the fluctuations caused by gestures are highly coupling with body features and the user's location, we propose a WiFi-based multi-task sensing model (Wimuse) to perform gesture recognition, indoor localization, and user identification tasks simultaneously. However, these tasks have different difficulty levels (i.e., imbalance issue) and need task-specific information (i.e., discrepancy issue). To address these issues, the knowledge distillation technique and task-specific residual adaptor are adopted in Wimuse. We first train the STS model for each task. Then, for solving the imbalance issue, the extracted common feature in Wimuse is encouraged to get close to the counterpart features of the STS models. Further, for each task, a task-specific residual adaptor is applied to extract the task-specific compensation feature which is fused with the common feature to address the discrepancy issue. We conduct comprehensive experiments on three public datasets and evaluation suggests that Wimuse achieves state-of-the-art performance with the average accuracy of 85.20%, 98.39%, and 98.725% on the joint task of gesture recognition, indoor localization, and user identification, respectively.

## 1 Introduction

WiFi-based sensing has drawn considerable interest over recent years due to its pervasive availability, non-intrusiveness, and low-cost deployment. Numerous studies [1]–[3] have shown that WiFi-based human sensing can be regarded as a promising candidate to promote human-computer interaction in the Internet of Things (IoT) era. The basic principle of WiFi-based sensing is that some of the signals are absorbed, reflected, or scattered by humans on the propagation, leading to the signal fluctuations which carry the information of human behavior. Further, the signal fluctuations are described by the channel state information (CSI) of WiFi, which can be captured from commercial WiFi devices [4], [5].

Despite significant advances in the field of WiFi-based sensing, various pioneering approaches are limited to single-task sensing (STS), such as human activity recognition [6], indoor localization [7], gait identification [8], breath detection [9]. There are some studies on WiFi-based multi-user gesture recognition [10], [11]. However, to our best knowledge, ARIL [1] and WiHF [12] are the only two works on WiFi-based multi-task sensing (MTS). Specifically, ARIL aims to perform the joint task of activity recognition and indoor localization, constructed on the universal software radio peripheral devices, not the commercial WiFi devices, which decreased the practicability. WiHF focuses on simultaneously enabling cross-domain gesture recognition and user identification, which can only deal with the joint gesture recognition and user identification task.

Based on the observation that the signal fluctuations are related to human gestures, body features, and indoor locations, we propose a WiFi-based multi-task sensing model (Wimuse) to perform gesture recognition, user identification, and indoor localization simultaneously. The exploitation of WiFi-based MTS will provide more users' information than STS, which will promote many IoT applications. For example, in the smart home application, Wimuse can provide the information of 'who does what and where?' that will facilitate the smart home system precisely responds to the user. More specifically, we can use 'hand up' to increase the television volume in front of the user and turn on the light with the same gesture in the bathroom. In addition, to ensure safety, adults can use 'draw circle' to turn on the stove, while children cannot use the same gesture to open it.

However, there are two issues in MTS. i) The difficulty levels of these tasks are different, leading to the imbalance issue [13]. Specifically, the gesture recognition task is more difficult than both the indoor localization and the user identification tasks, which may cause two types of undesirable situations in the training phase: the performance of the indoor localization and the user identification tasks are far superior to that of the gesture recognition task. In another case, the gesture recognition task performed well, while the other two tasks are overfitting. ii) Different tasks need task-specific information, namely, the task discrepancy issue. For example, the gesture recognition task needs more detailed pose information than the indoor localization task. However, the detailed pose information is redundant information for the indoor localization task, while this task requires more information about the spatial distribution of objects. Even worse, the useful information of one task may be others' noise [14].

To address these issues, we adopt the knowledge distillation technique [15] and task-specific residual adaptor [16] in Wimuse. Concretely, inspired by the knowledge-distillation-based method [13], we first train the STS model for each task which provides the task-specific feature. Then, in the training phase, we encourage the extracted common feature of Wimuse to get close to all the task-specific features after a linear transform under the Euclidean distance. In this way, the common feature will prevent being dominated by a particular task. For the discrepancy issue, a task-specific residual adaptor is added to extract the task-specific compensation feature for each task which concatenates with the common feature to form the feature of each task in Wimuse. In addition, the predictive logits for each task in Wimuse are encouraged to be similar to that produced by the corresponding trained STS model, which will transfer their generalization ability to Wimuse.

We conduct comprehensive experiments on three public datasets: ARIL dataset [1], CSIDA dataset [17], and Widar3.0 dataset [18]. The experimental results show that Wimuse achieves state-of-art performance on the joint gesture recognition, localization, and user identification task. In summary, our contributions are listed as follows.

i.  To our best knowledge, Wimuse is the first approach of WiFi-based MTS involving more than two tasks and achieve promising performance.
ii. We reveal the imbalance and discrepancy issues in WiFi-based MTS and novelly adopt the knowledge distillation technique and task-specific residual adaptor to address these issues.
iii. We evaluate Wimuse on three public datasets, the results show that Wimuse achieves the average accuracy of 85.20%, 98.39%, and 98.725% for the joint gesture recognition, indoor localization, and user identification tasks.

## 2    Related Work

In this section, we first introduce the WiFi-based sensing approaches. Then, some studies on multi-task learning and knowledge distillation are briefly presented.

### 2.1    WiFi-based Sensing

Numerous studies of WiFi-based sensing have been proposed on many applications, including gesture recognition, indoor localization, and human identification, etc.

In [3], the authors proposed SignFi, based on a convolutional neural network (CNN), to recognize gestures. The experimental results showed the average accuracy is 86.66% by using the CSI as input, which shown the effectiveness of CNN for extracting the features from CSI samples. To promote the practicability of WiFi-based sensing, [19] proposed a WiFi-based gesture recognition system for smartphones. In addition, due to the environmental sensitivity of WiFi signals, [20] adopted adversarial learning to train environment-invariant feature extractors to construct a robust WiFi-based gesture recognition system. It is worth noting that [10] and [11] accomplished gesture recognition systems that can recognize multiple users' gestures simultaneously.

WiFi-based indoor localization approaches can be summarized into two categories: **i) Physical approaches** try to estimate physical factors of WiFi signal, such as Angle of Arrival (AOA), Time of Flight (TOF), and Doppler Frequency Shift (DFS). Then, the target position is calculated through physical and geometric models based on the estimated physical factors. For example, [21] proposed an indoor localization system, named Phaser, based on the estimation of AOA. In addition, K. Wu et al. [22] proposed a fine-grand indoor localization method FILA to directly estimate the distance between the object and the three fixed access points. **ii) Pattern-based**

**approaches** are dedicated to finding the location-discriminate representation of CSI to perform the localization task. Fingerprint-based localization is the most commonly used pattern-based approach, the main idea of which is to collect CSI samples of all possible locations in the area of interest to build a fingerprint database. Next, two methods are applied to perform the localization using the fingerprints. One is to match the unknown fingerprints with those in the database and return the location of the best-fitted fingerprint [7]. The other is to train a deep learning model to map the fingerprints to their location labels using the fingerprint database as a training dataset, then return the location label of the trained model with the unknown fingerprint as input [1], [23], [24]. Note that, in this work, we adopt the latter method for the indoor localization task.

User identification is the process of associating a user with a predefined identity. WiHF [12] performed user identification and gesture recognition using CSI in a real-time manner based on the difference between the personal user performing styles and the unique gesture characteristics. In [25], the authors conducted a feasibility study for user identification using the CSI of the user's in-air handwritten signatures. In this work, the user identification is performed based on the observation that the CSI is different with different users even at the same position and performing the same gesture.

## 2.2   Multi-task Learning

Multi-task learning is a subfield of machine learning that is dedicated to constructing one model to perform multiple tasks simultaneously. Multi-task learning has been studied in various research fields, including computer vision [26], natural language processing [27], speech recognition [28], etc. However, according to existing works, two challenges hinder the better performance and efficiency to be improved.

The first one is called task-imbalance that the multi-task learning model is dominated by a particular task leading to unacceptable performances of the other tasks. Some previous studies addressed this issue with balanced loss weighting and parameter updating strategies [29], [30]. In [31], the authors explicitly cast multi-task learning as multi-objective optimization, with the overall objective of finding a Pareto optimal solution. Wei-Hong Li et al. [13] constructed a novel idea combining the knowledge distillation method to solve the task imbalance issue. Based on this idea, a more in-depth study is conducted and some new solutions are proposed in this paper.

The other is task-discrepancy for different tasks may require different task-specific information. The lack of task-specific information may degrade the performance of the multi-task learning model. Even worse, the task-specific information of different tasks may contract with each other, which means that one's useful information may others' noise [14]. In this work, we adopt the residual adaptor blocks [16] to extract the task-specific compensation information to address this issue.

### 2.3 Knowledge Distillation

Knowledge distillation (KD), proposed by Hinton et al. [15], is dedicated to transforming the learned knowledge in an ensemble of models into a single model and is adopted in model compression [32], transfer learning [33], domain adaptation [34], and multimodal learning [35], etc.

In addition to success in single-task learning, knowledge distillation has also been proved to be efficient in multi-task learning. Dan Xu et al. [36] proposed a novel multi-task guided prediction-and-distillation network to perform the joint tasks for depth estimation and scene parsing. Sahil Chelaramani et al. [37] proposed the use of multi-task learning and knowledge distillation to improve fine-grained recognition of eye diseases using a small labeled dataset of fundus images.

In this work, we adopt the knowledge distillation technique to address the task-imbalance issue, which is inspired by the knowledge distillation-based method [13]. In addition, to sufficiently exploit the trained STS model, we add a new loss in Wimuse to encourage the predictive logits for each task to get closer to that produced by the corresponding trained STS model. The experimental results conduct that this trick can promote the performance of Wimuse.

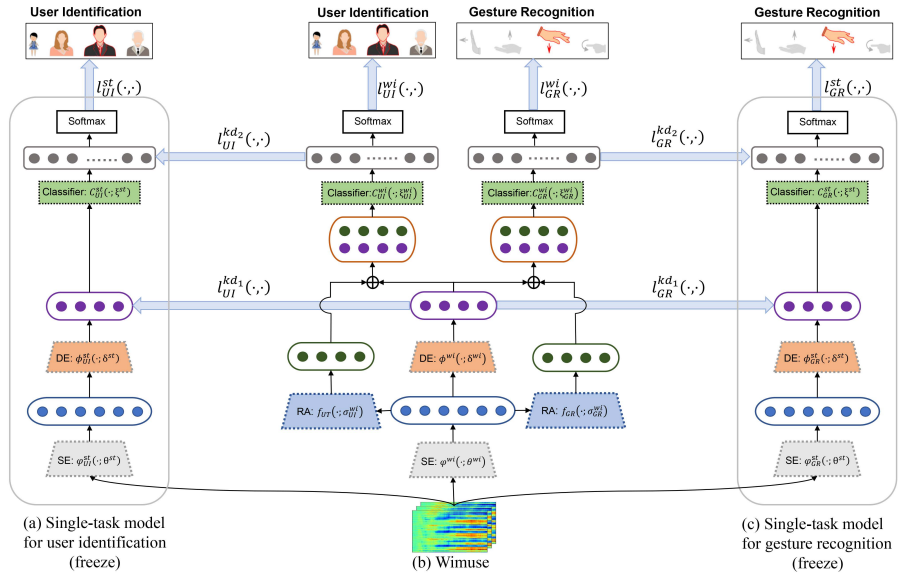

**Fig. 1.** The framework of Wiumuse (an example in the two-task condition, i.e. user identification and gesture recognition). We first train the task-specific STS model for each task (i.e. (a) and (c)). Then, freezing the learned parameters of the STS models. In the training phase of Wimuse, we optimize the parameters by minimizing the losses.

# 3 Methodology

In this section, we present the framework of Wimuse as shown in Fig. 1. Concretely, we first introduce the input and four basic modules (i.e., shallow encoder, deep encoder, task-specific residual adaptor, and classifier). Then, we will discuss the STS model and the naïve MTS model. Finally, Wimuse is proposed.

## 3.1 Overview

As illustrated in Fig.1, we adopt the CSI sample as input to implement the sensing tasks. Specifically, CSI reflects multipath fluctuations of WiFi signal on all subcarriers. For each subcarrier, we can model the channel impulse response as:

$$h(t) = \sum_{i=0}^{I-1} a_i \delta(t - \tau_i) e^{-j2\pi f_i} \quad (1)$$

where $I$ denotes the number of signal paths. $a_i$, $2\pi f_i$, and $\tau_i$ are the attenuation factor, phase shift, and time delay of the signal on the $i$-th path, respectively. In addition, $f_i$ represents the frequency of subcarrier, $\delta(t)$ is the Dirac delta function.

Further, the orthogonal frequency division multiplexing receiver can sample the CIR in all subcarriers to construct a complex matrix, i.e., a CSI matrix. Hence that, CSI can be denoted as:

$$H = \|H\| e^{j\angle H} \quad (2)$$

where $H$ denotes the CSI measurement on all subcarriers. $\|H\|$ and $\angle H$ represent the amplitude and phase, respectively.

In this work, we adopt $\|H\| \in R^{L \times S \times T}$ as input, where $L$ is the number of links, namely, $L = RX \times TX$, $RX$ and $TX$ is the number of antennas of the receiver and the transmitter. $S$ denotes the number of subcarriers. $P$ represents the length of sampling, which is determined by the sampling rate $r$ of the CSI capture tool and the gesture execution time t (i.e., $P = r \times t$). hence that, the input $\|H\| \in R^{L \times S \times P}$ can be regarded as the reflection of signal fluctuations on three dimensions, i.e., spatial, frequency, and time.

In addition, we provide four basic modules to construct the STS model, the naïve MTS model, and Wimuse. Note that the implementation details of these modules are presented in Section IV, while a brief introduction is as follows.

i. **shallow encoder (SE)**: For extracting the features on links of input $\|H\| \in R^{L \times S \times P}$, the shallow encoder, denoted as $\varphi(\cdot)$, contains a grouping convolutional layer and a max-pooling layer, where the number of groups equals the number of links $L$.

ii. **deep encoder (DE)**: By using the low-level features extracted by the Shallow Encoder as input, a deeper convolutional subnetwork $\phi(\cdot)$ is adopted to fusion the features of different links and extract high-level features. In addition, this module is constructed based on the structure of ResNet [38].

iii. **task-specific residual adaptor (RA)**: To deal with the task discrepancy issue, a Task-specific residual adaptor $f(\cdot)$, which contains only one convolutional layer, is adopted in the Wimuse to gain the task-specific compensation feature from the low-level features.

iv. **classifier**: For each classification task, a classifier $C(\cdot)$, which contains a convolutional layer and a fully connected layer, is adopted to output the logits. Then, by using the softmax function on the predictive logits, the probability over categories is produced.

Note that, all the convolutional layers in the above modules adopt one-dimensional convolution along the time dimension of CSI. We using superscripts $st$, $mt$, and $wi$ to denote the modules in the STS model, the naïve MTS model, and the Wimuse model, respectively. In addition, the subscript of the above modules' symbols represents the task, i.e., $UI$ denotes user identification, $GR$ is the gesture recognition, and $IL$ presents indoor localization. For example, $C_{UI}^{st}(\cdot)$ presents the Classifier module of an STS model at the user identification task.

### 3.2 Single-task Sensing Model

Consider that we have a dataset D that contains N training CSI samples $H_n$ and their labels $y_n \in \{1,2,\dots,M\}$, for n $\in \{1,2,\dots,N\}$, where M is the number of categories. In this case, the STS model is composed of three parts: **i)** a shallow encoder $\varphi^{st}(\cdot;\theta^{st})$, where $\theta^{st}$ are the learned parameters. Given a CSI sample $H_n$ as input, the output of $\varphi^{st}(\cdot;\theta^{st})$ is the low-level features on all links, i.e. $F_n^{st,low} = \varphi^{st}(H_n;\theta^{st})$. **ii)** a deep encoder $\phi^{st}(\cdot;\delta^{st})$ is adopted to fusion the low-level features across links and map to high-level features $F^{st,high}$, where $\delta^{st}$ denotes the learned parameters. Hence, the output of the Deep Encoder is $F_n^{st,high} = \phi^{st}(\cdot;\delta^{st}) \circ \varphi^{st}(H_n;\theta^{st})$. **iii)** To output the predictive probability distribution over categories, a classifier $C(\cdot;\xi^{st})$ is used to output the predictive logits $\mathbf{Z}^{st} \in R^M$, where $\xi^{st}$ presents the learned parameters. Particularly, the predictive logits of $H_n$ is $\mathbf{Z}_n^{st} = C(\cdot;\xi^{st}) \circ \phi^{st}(\cdot;\delta^{st}) \circ \varphi^{st}(H_n;\theta^{st})$. Then, the predictive probability distribution $p_n$ of $H_n$ over categories is produced by using the softmax function as follows:

$$p_n(\widehat{y_n} = m|H_n) = \sigma(\mathbf{Z}_n^{st})_m \tag{3}$$

where $\widehat{y_n}$ is the predicted label of $H_n$ and $m \in \{1,2,\dots,M\}$. $\sigma(\cdot)_m$ is the softmax function, defined as:

$$\sigma(\mathbf{Z})_m = \frac{e^{Z_m}}{\sum_{k=1}^{M} e^{Z_k}}, for\ \mathbf{Z} \in R^M, m \in \{1,2,\dots,M\} \tag{4}$$

To optimize the parameters of the STS model, a loss function $l^{st}(\cdot,\cdot)$ based on cross-entropy is adopted. The loss for $H_n$ is as follows:

$$l^{st}(p_n, y_n) = -\sum_{m=1}^{M}(1|y_n = m))\log(p_n(\widehat{y_n} = m|H_n)) \tag{5}$$

where $(1|y_n = m))$ means if $y_n = m$ is true, $(1|y_n = m))$ equals 1, otherwise, 0.

The goal is to minimize the sum of losses overall training samples.

$$(\theta_{opt}^{st}, \delta_{opt}^{st}, \xi_{opt}^{st}) = arg \min_{\theta^{st},\delta^{st},\xi^{st}} \sum_{n=1}^{N} l^{st}(p_n, y_n) \tag{6}$$

### 3.3 Naïve Multi-task Sensing Model

The naïve MTS model is based on the classic multi-task learning model proposed in [26]. Specifically, the naïve MTS model has only one difference from the STS model. Consider that we are given a dataset D containing N training CSI samples $H_n$ and their labels $\boldsymbol{y_n}$ where n ∈ {1,2,...,N}. Note that, there are T tasks to performance, i.e. $\boldsymbol{y_n} \in R^T$. The element $y_{t,n} \in \{1,2,...,M_t\}$ is the label of $H_n$ on the t−th task, where $M_t$ is the number of categories on the t-th task.

The naïve MTS model has the same encoder structure as the above STS model. Then the high-level features $F_n^{mt,high}$ of $H_n$ is $\phi^{mt}(\cdot;\delta^{mt}) \circ \varphi^{mt}(H_n;\theta^{mt})$. Compared with the STS model, there are T classifiers in the naïve MTS model, e.g. $C_t^{mt}(\cdot;\xi^{mt})$ is the classifier of the t-th task. Further, the predictive logits of $H_n$ on the t -th task is $\boldsymbol{Z_{t,n}^{mt}} = C_t^{mt}(\cdot;\xi_t^{mt}) \circ \phi^{mt}(\cdot;\delta^{mt}) \circ \varphi^{mt}(H_n;\theta^{mt}) \in R^{M_t}$. Finally, the predictive probability distribution of $H_n$ over category on the t-th task is:

$$p_{t,n}(\widehat{y_{t,n}} = m_t|H_n) = \sigma(\boldsymbol{Z_{t,n}^{mt}})_{m_t}, for\ m_t \in \{1,2,...,M_t\} \qquad (7)$$

where $\widehat{y_{t,n}}$ is the predicted label of $H_n$ on the t-th task, $m_t \in \{1,2,...,M_t\}$. $\sigma(\cdot)_{m_t}$ is the softmax function as (4).

The loss function for the t-th task $l_t^{mt}(\cdot,\cdot)$ in naïve MTS model is the same as that in STS model, i.e. $l_t^{mt}(p_{t,n}, y_{t,n}) = l^{st}(p_{t,n}, y_{t,n})$. Then the linear combination of all the task-specific loss constructs the final loss of the naïve MTS model, namely,

$$l^{mt}(\boldsymbol{p_n}, \boldsymbol{y_n}) = \sum_{t=1}^{T} \omega_t l_t^{mt}(p_{t,n}, y_{t,n}) \qquad (8)$$

where $\omega_t$ is a hyperparameter for the t-th task, $\boldsymbol{p_n} = \{p_{1,n}, p_{2,n}, \cdots, p_{T,n}\}$ is the set of predictive probability distributions of the sample $H_n$ of all tasks.

Finally, the naïve MTS model can be learned by optimizing the following loss:

$$(\theta^{mt}, \delta^{mt}, \xi_1^{mt}, \xi_2^{mt}, \cdots, \xi_T^{mt})_{opt} = arg \min_{\theta^{mt}, \delta^{mt}, \xi_1^{mt}, \xi_2^{mt}, \cdots, \xi_T^{mt}} \sum_{n=1}^{N} \sum_{t=1}^{T} \omega_t l_t^{mt}(p_{t,n}, y_{t,n}) \qquad (9)$$

### 3.4 Wimuse Model

The Wimuse model is based on the naïve MTS model while introducing the knowledge distillation method and task-specific residual adaptor to address the task imbalance and discrepancy issues.

Given the same dataset D as the above MTS model. In this case, the Wimuse model is composed as following parts:

**i)** We first adopt a shallow encoder, denoted as $\varphi^{wi}(\cdot;\theta^{wi})$, to extract the low-level features on all links, $\theta^{wi}$ is the learned parameters. For example, the low-level features of the sample $H_n$ is $F_n^{wi,low} = \varphi^{wi}(H_n;\theta^{wi})$.

**ii)** To address the task discrepancy issue, we adopt a task-specific residual adaptor $f(\cdot;\sigma^{wi})$ for each task to extract the task-specific compensation feature, denoted as $F^{wi,comp}$. For example, the task-specific compensation feature of $H_n$ at the t − th task is $F_{t,n}^{wi,comp} = f_t(\cdot;\sigma_t^{wi}) \circ \varphi^{wi}(H_n;\theta^{wi})$.

**iii)** The common feature $F_n^{wi,comm}$ of sample $H_n$ for all tasks are extracted by a deep encoder $\phi^{wi}(\cdot;\delta^{wi})$ using $F_n^{wi,low}$ as input, i.e. $F_n^{wi,comm} = \phi^{wi}(F_n^{wi,low};\delta^{wi})$.

**iv)** Similar to the MTS model, Wimuse also have T classifiers, e.g. for the t − th task, the classifier is $C_t^{wi}(\cdot;\xi_t^{wi})$. Note that, the input for each classifier is the composition of the corresponding task-specific compensation feature and the common features. For example, the input for $C_t^{wi}(\cdot;\xi_t^{wi})$ of the sample $H_n$ is $F_{t,n}^{wi,comp} \oplus F_n^{wi,comm}$, where $\oplus$ presents the concatenate operation. In addition, the output of $C_t^{wi}(\cdot;\xi_t^{wi})$, for the sample $H_n$, is the predictive logits of the t-th task, i.e.

$$\mathbf{Z}_{t,n}^{wi} = C_t^{wi}(F_{t,n}^{wi,comp} \oplus F_n^{wi,comm};\xi_t^{wi}) \tag{10}$$

For the task imbalance issue, a knowledge-distillation-based method [13] is adopted in Wimuse. Concretely, for each task, we first train an STS model, e.g. for the t-th task, we train an STS model $M_t^{st}(\cdot;\theta^{st},\delta^{st},\xi^{st})$. Then we freeze the learned parameters of this model. Further, with an input $H_n$, $M_t^{st}$ can provide the high-level features $F_{t,n}^{st,high}$ and the logits $\mathbf{Z}_{t,n}^{st}$ for the t-th task. Then, at the training phase of Wimuse, we construct a loss function $l_t^{kd_1}(\cdot,\cdot)$ to encourage the common feature $F_n^{wi,comm}$, extracted by Wimuse, to get close to all of the high-level features produced by the trained STL models after a linear transform under the Euclidean distance. For example, for the t-th task on the sample $H_n$, the loss $l_{t,n}^{kd_1}$ is as follows:

$$l_{t,n}^{kd_1} = \left\| \frac{LT_t(F_n^{wi,comm})}{\|LT_t(F_n^{wi,comm})\|_2} - \frac{F_{t,n}^{st,high}}{\|F_{t,n}^{st,high}\|_2} \right\|_2 \tag{11}$$

where $LT_t(\cdot)$ is the linear transform for the t-th task, which is implemented as a $1 \times 1 \times C \times C$ convolution, where $C$ is the depth (number of channels) of $F_n^{wi,comm}$.

Furthermore, to sufficiently exploit the trained STS models, we also encourage the predictive logits $\mathbf{Z}_{t,n}^{wi}$ to be similar with the logits $\mathbf{Z}_{t,n}^{st}$ for sample $H_n$ on the t-th task, which leads to another loss:

$$l_{t,n}^{kd_2} = -\sum_{m_t=1}^{M_t} \left(P_{1,m_t} \cdot \log(P_{2,m_t})\right) \tag{12}$$

Where $P_{1,m_t}$ and $P_{2,m_t}$ is the predictive probability distribution based on $\mathbf{Z}_{t,n}^{wi}$ and $\mathbf{Z}_{t,n}^{st}$, calculated as follows:

$$P_{1,m_t} = \sigma\left(\frac{\mathbf{Z}_{t,n}^{wi}}{\tau}\right)_{m_t} \tag{13}$$

$$P_{2,m_t} = \sigma\left(\frac{\mathbf{Z}_{t,n}^{st}}{\tau}\right)_{m_t} \tag{14}$$

Where $\sigma(\cdot)_{m_t}$ is the softmax function as (4). And $\tau$ is a hyperparameter to adjust the intensity of distillation [15].

The final output of Wimuse for input sample $H_n$ is a set of the probability distribution $\mathbf{p}_n =$

$\{p_{1,n}, p_{2,n}, \cdots, p_{T,n}\}$ of all T tasks. The element is defined as follows:
$$p_{t,n}(\widehat{y_{t,n}} = m_t | H_n) = \sigma(Z_{t,n}^{wi})_{m_t} \tag{15}$$
where $\widehat{y_{t,n}}$ is the predicted label of $H_n$ on the t-th task, and $m_t \in \{1,2,\ldots,M_t\}$. $\sigma(\cdot)_{m_t}$ is the softmax function as (4).

The loss function for the final prediction is as follows,
$$l^{wi}(\boldsymbol{p_n}, \boldsymbol{y_n}) = \sum_{t=1}^{T} \omega_t l_t^{wi}(p_{t,n}, y_{t,n}) \tag{16}$$
where $\omega_t$ is a hyperparameter for the t-th task and $l_t^{wi}(\cdot,\cdot)$ is identical with $l^{st}(\cdot,\cdot)$ (refer to (5)).

The parameters of Wimuse can be learned by minimizing the following loss:
$$l(D) = \sum_{n=1}^{N} \sum_{t=1}^{T} \left( \omega_t l_t^{wi}(p_{t,n}, y_{t,n}) + \lambda l_{t,n}^{kd_1} + l_{t,n}^{kd_2} \right) \tag{17}$$
where D is the training set and $\lambda$ is the weight of the loss $l_t^{kd_1}$.

---

**Algorithm 1**: Episode-based training for Wimuse

$\boldsymbol{\varphi}_t^{st}(\cdot)$: the pre-trained shallow encoder of single task model (STM) for t-th task.
$\boldsymbol{\phi}_t^{st}(\cdot)$: the pre-trained deep encoder of STM for t-th task.
$\boldsymbol{C}_t^{st}(\cdot)$: the pre-trained classifier of STM for t-th task.
$\boldsymbol{\varphi}^{wi}(\cdot)$: the shallow encoder of Wimuse.
$\boldsymbol{\phi}^{wi}(\cdot)$: the deep encoder of Wimuse.
$\boldsymbol{C}_t^{wi}(\cdot)$: the classifier of the t-th task in Wimuse.
$\boldsymbol{f}_t(\cdot)$: the task-specific residual adaptor of Wimuse for the t-th task.

**Input:** Training set P= $\{(x_i, y_{t,i})\}_{i=1}^{N}$, where $y_{t,i} \in \{1,2,\ldots,M_t\}$ is the label of sample $x_i$ on the $t-$th task, and where $M_t$ is the number of categories on the t-th task.
**Output:** The loss J of a randomly generated episode.

**Beginning:**
    Give a sample $(x_i, y_{t,i})$ from P randomly.
    $\boldsymbol{F}^{wi,comm} = \boldsymbol{\phi}^{wi}(\cdot) \circ \boldsymbol{\varphi}^{wi}(x_i)$.
    // calculate the common feature in Wimuse
    $\boldsymbol{F}_{t,i}^{wi,comp} = \boldsymbol{f}_t(\cdot) \circ \boldsymbol{\varphi}^{wi}(x_i)$
    // calculate the compensation feature of the t-th task in Wimuse
    for t in {1,2, ... T} do
        $Z_t^{wi} = C_t^{wi}(F_{t,n}^{wi,comp} \oplus F_n^{wi,comm})$
        // calculate the predictive logits of the t-th task in Wimuse
        $F_t^{st,high} = \boldsymbol{\phi}_t^{st}(\cdot) \circ \boldsymbol{\varphi}_t^{st}(x_i)$
        // calculate the high-level feature of the t-th task in pre-trained STM.
        $Z_t^{st} = C_t^{st}(\cdot) \circ \boldsymbol{\phi}_t^{st}(\cdot) \circ \boldsymbol{\varphi}_t^{st}(x_i)$.
        // calculate the predictive logits of the t-th task in pre-trained STM.
        Calculate $\boldsymbol{l}_{t,i}^{kd_1}$ (where the kd1 loss of the t-th task) by using (11).

```
            Calculate l_{t,i}^{kd2} (i.e., the kd2 loss of the t-th task) by using (12).
            Calculate l_{t,i}^{wi} (i.e., the final loss of the t-th task) by using (16).
    End for
    L ← 0                                                            // Initialize loss.
    L ← ∑_{t=1}^{T} (ω_t l_t^{wi}(p_{t,n}, y_{t,n}) + λ l_{t,n}^{kd1} + l_{t,n}^{kd2})   // Update loss.
End
```

## 4 Experiment

In this section, we evaluate Wimuse on three public datasets under two-task and three-task sensing scenarios through using the amplitude data. As for the whole dataset, we divide 80% of them for training and 20% for testing.

CSI measured on commercial WiFi devices is well-known to contain phase offsets, including carrier frequency offset (CFO), sampling frequency offset (SFO), and symbol timing offset (STO). These phase offsets are strongly relevant to the CSI capture device, which means the phase data of various public datasets has different phase offsets and we can not use the same methods to eliminate them. In addition, we are inclined to have a more general model. Therefore, we give up the phase data of these public datasets and just use the amplitude data of them.

The pseudocode of the training scheme of Wimuse is in Algorithm 1, and the python code of Wimuse is available at https://github.com/Zhang-xie/Wimuse.

### 4.1 Datasets

**ARIL dataset.** ARIL dataset is proposed in [1] for the joint task of activity recognition and indoor location. Specifically, it contains the CSI samples of six gestures (i.e., up, down, left, right, circle, and cross.) in 16 locations of one room by one volunteer. In addition, the CSI samples were collected by using a pair of universal software radio peripherals (USRPs) and the total number of samples is 1440. Each sample has the shape of $1 \times 52 \times 192$, i.e. one link, 52 subcarriers, and 192 packets for one sample.

**CSIDA dataset.** CSIDA dataset [17] contains CSI samples of six gestures (i.e., hand left, hand right, lift, press, draw circle, and draw zigzag) in five different locations by five users. The samples are captured from laptops (i.e., one transmitter and one receiver) equipped with the Atheros CSI tool [39]. In addition, devices are set to work at monitor mode at 5 GHz to capture information of 114 subcarriers. The transmitter and receiver activate one and three antennas, respectively. Further, the sampling rate is 1000 packets per second and the gesture execution time is 1.8 seconds. There are 1500 samples with the shape $3 \times 114 \times 1800$.

**Widar3.0 dataset.** Widar3.0 [18] contains two sub-datasets: Dataset1 contains 12000 CSI samples collected from 16 users performing six gestures (push & pull, sweep, clap, slide, draw a circle and draw zigzag) in five different locations at three different environments, i.e., a classroom, an office, and a hall. Dataset2 holds 5,000 instances of two volunteers (one male and one female) drawing numbers 0-9 in a horizontal plane. In addition, each CSI sample in the Widar3.0 dataset is collected by six receivers and one transmitter all with three antennas. Due to the use of the 802.11n CSI tool [5], there are 30 subcarriers for each link, and send 1000 package per second. The CSI sample of one receiver has the shape of $3 \times 30 \times P$, where $P$ represents the length of sampling from 1300 to 2200. In this work, we only adopt the samples from one receiver and Dataset1 to evaluate gesture recognition, user identification, and localization tasks.

**Table 1.** The architectures of the basic modules

| Module | Layers | Input (No. channel × length) | Kernel size | stride |
|---|---|---|---|---|
| shallow encoder | 1D-Convolution | $(L*S) \times P$ | $7 \times 1$ | 2 |
| | 1D-BatchNorm | $(128*L) \times (P/2)$ | -- | -- |
| | 1D-Max-pooling | $(128*L) \times (P/2)$ | $3 \times 1$ | 2 |
| deep encoder | 1D-Convolution | $(128*L) \times (P/4)$ | $3 \times 1$ | 1 |
| | 1D-BatchNorm | $(128*L) \times (P/4)$ | -- | -- |
| | 1D-Convolution | $(128*L) \times (P/4)$ | $3 \times 1$ | 1 |
| | 1D-BatchNorm | $(128*L) \times (P/4)$ | -- | -- |
| | 1D-Convolution | $(128*L) \times (P/4)$ | $3 \times 1$ | 2 |
| | 1D-BatchNorm | $(128*L) \times (P/8)$ | -- | -- |
| | 1D-Convolution | $(128*L) \times (P/8)$ | $3 \times 1$ | 1 |
| | 1D-BatchNorm | $(128*L) \times (P/8)$ | -- | -- |
| | 1D-Convolution | $(128*L) \times (P/8)$ | $3 \times 1$ | 2 |
| | 1D-BatchNorm | $(128*L) \times (P/8)$ | -- | -- |
| | 1D-Convolution | $(128*L) \times (P/8)$ | $3 \times 1$ | 1 |
| | 1D-BatchNorm | $(128*L) \times (P/8)$ | -- | -- |
| Classifier | 1D-Convolution | $(128*L) \times (P/8)$ | $3 \times 1$ | 1 |
| | 1D-BatchNorm | $(128*L*2) \times 10$ | -- | -- |
| | 1D-AadAvgPool | $(128*L*2) \times 10$ | -- | -- |
| | Linear | $(128*L*2) \times 1$ | -- | -- |
| residual adaptor | 1D-Convolution | $(128*L) \times (P/4)$ | $3 \times 1$ | 2 |
| | 1D-BatchNorm | $(128*L) \times (P/8)$ | -- | -- |

$S$ denotes the number of CSI subcarriers. $P$ represents the time length of the CSI sample. $L$ is the number of links. 1D is the abbreviation of one-dimensional. AadAvgPool is the adaptive average pooling layer.

### 4.2 Baselines

In this work, we compare our method Wimuse with four baselines. Note that, these baselines are constructed for CSI samples with the proposed basic modules keeping the original framework. The architecture details of the basic modules are illustrated in Table 1.

- The **STS** model for each task.
- **NMTS**: The naïve MTS model learned by minimizing the loss in (9) with the hyperparameters $\omega_t = 1$ for all tasks.
- **UMTS:** The naïve MTS model is enhanced by a principled approach [29] which weighs multiple loss functions by considering the homoscedastic uncertainty of each task to address the task unbalanced issue.
- **KDMTS:** This approach [13] addresses the unbalanced issue based on knowledge distillation. Since this method is published for joint image segmentation and depth estimation tasks, we implement this method with the abovementioned basic modules keeping the original framework.

### 4.3 Comparison

We compare our method to the baselines. Besides, on the ARIL dataset, we also compared it with the ARIL method [1]. We use the mean accuracy as an evaluation standard.

- **Table 2.** Performance of Various Methods on ARIL dataset (Accuracy: %)

| Type | Method | Two-task Sensing GR | IL | Average |
|---|---|---|---|---|
| STS | -- | 88.77 | 98.69 | 93.73 |
|  | NMTS | 93.19 | 98.45 | 95.82 |
|  | ARIL | 88.13 | 95.68 | 91.91 |
| MTS | UMTS | 94.15 | 98.80 | 96.48 |
|  | KDMTS | 93.79 | 98.45 | 96.12 |
|  | Wimuse (ours) | **95.70** | **99.16** | **97.43** |

GR, IL are the abbreviations of gesture recognition and indoor localization, respectively.

**Results on ARIL.** Firstly, we evaluate all methods on the ARIL dataset. We set the minibatch size as eight and use Adam [40] for optimizing the models. The initial learning rate is set to 0.001 and we train all methods for 500 epochs in total where we scaled the learning rate by 0.5 every 100 epochs until the 350th epoch. In Wimuse, weights of task-specific residual adaptor losses (i.e. E. (11)) and logits losses (i.e. E. (12)) are set uniformly. After the hyperparameter (λ) search, it is set 4.0 for KDMTS, and 8.0 for Wimuse. In addition, another hyperparameter (τ) to adjust the intensity of distillation is also set to 8.0 in Wimuse.

As shown in Table 2, except for the ARIL method, the MTS methods obtain better overall performance than STS, which proves that MTS models can learn more informative features than the STS model. Furthermore, ARIL obtains worse overall performance because of its deep neural network and negligence of the abovementioned task-imbalance even task-discrepancy problem. In contrast, our method achieves significantly better performance than any other MTS method, i.e. our method achieves an accuracy of 95.70% for gesture recognition and 99.16% for indoor localization, which results from our solutions of the task-imbalance and task-discrepancy problem.

Table 3. Performance of Various Methods on Widar3.0 (Accuracy: %)

| Type | Method | Two-task Sensing | | Two-task Sensing | | Two-task Sensing | | Three-task Sensing | | |
|------|--------|------|------|------|------|------|------|------|------|------|
|      |        | GR   | IL   | GR   | UI   | IL   | UI   | GR   | IL   | UI   |
| STS  | --     | --   | --   | --   | --   | --   | --   | 80.42 | **96.96** | 97.19 |
| MTS  | NMTS   | 81.79 | 96.15 | 81.63 | 97.29 | 96.24 | 97.17 | 81.74 | 96.38 | 96.78 |
|      | UMTS   | 80.7 | 96.19 | 82.09 | 97.45 | 95.36 | 97.26 | 81.70 | 96.38 | 96.78 |
|      | KDMTS  | 82.53 | 96.24 | 81.44 | 97.17 | 96.20 | 97.52 | 83.65 | 96.92 | 97.01 |
|      | Wimuse (ours) | **85.27** | **96.84** | **82.46** | **97.59** | **96.46** | **97.91** | **83.79** | 96.73 | **98.05** |

GR, IL, and UI are the abbreviations of gesture recognition, indoor localization, and user identification.

**Results on Widar3.0 dataset.** Different from the experimental setup on ARIL, we performed two groups of experiments on the Widar3.0 dataset. 1) the two-task sensing including the joint gesture recognition and indoor localization task, the joint gesture recognition and user identification task, and the joint indoor localization and user identification task. 2) The three-task sensing is the joint of gesture recognition, IL, and user identification tasks. Corresponding to the four experiments, we set the hyperparameter λ and τ to the same value 2.0. Beyond this, the other experimental setups are the same as those in experiments on the ARIL dataset.

From the results shown in Table 3, we can see that it is possible to tackle multiple sensing tasks within a network and achieve performance improvement on some tasks, e.g. compared with STS on the three-task sensing experiment, the NMTS model achieves better performance on gesture recognition though it causes indoor localization and user identification to a tiny drop. Though the effectivities of using NMTS, it is also clear that the task-imbalance even task-discrepancy problem exists for the recognition accuracy of gesture recognition is lower a lot than both of that in indoor localization and user identification tasks. Then we apply existing methods for solving the task-imbalance problem. From the results of using UMTS and KDMTS, UMTS performs as well as KDMTS obtains better overall performance than NMTS. However, they perform solely better on gesture recognition but worse on indoor localization and user identification in the comparison with the STS model.

In comparison with these methods, our method obtains significant overall performance and achieves better results than other MTS models and STS, which strongly verifies the effectiveness of our proposed strategies for feature-logits distillation and compensation of the task-specific features.

**Table 4.** Performance of Various Methods on CSIDA (Accuracy: %)

| Type | Method | Two-task Sensing | | Two-task Sensing | | Two-task Sensing | | Three-task Sensing | | |
|---|---|---|---|---|---|---|---|---|---|---|
| | | GR | IL | GR | UI | IL | UI | GR | IL | UI |
| STS | -- | -- | -- | -- | -- | -- | -- | 80.26 | 99.64 | 99.82 |
| MTS | NMTS | 82.31 | 99.76 | 83.24 | 99.88 | 99.82 | **99.97** | 84.30 | 99.82 | 99.40 |
| | UMTS | 70.12 | 99.94 | 80.02 | **99.98** | 99.88 | 99.88 | 78.38 | 99.76 | 99.82 |
| | KDMTS | 81.78 | 99.94 | 83.13 | 99.88 | 99.92 | 99.76 | 82.72 | 99.88 | 99.94 |
| | Wimuse (ours) | **83.07** | **99.94** | **83.48** | 99.24 | **99.92** | 99.82 | **84.53** | **99.94** | **99.97** |

GR, IL, and UI are the abbreviations of gesture recognition, indoor localization, and user identification.

**Results on CSIDA.** Similar to Widar 3.0, we set the hyperparameter $\lambda$ and $\tau$ to the same value for four different experiments, respectively $\lambda = 2.0$ and $\tau = 2.0$, as well.

As shown in Table 4, except for UMTS, the MTS models obtain better performance on gesture recognition tasks than the STS model and almost as excellent performance on indoor localization and user identification as the STS model. We assume that the UMTS may merely fit for applications in computer vision because its performance worse than NMTS. In addition, KDMTS performs slightly worse than NMTS, and our method remains outstanding performance among these methods.

### 4.4 Discussion

To better analyze the effect of the task-specific residual adaptor and the logits distillation, we conduct an ablation study on three datasets.

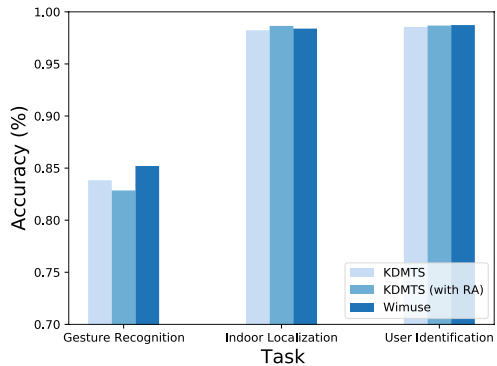

**Fig. 2.** The effects of the task-specific residual adaptor and the logits distillation

**Analysis of task-specific residual adaptor.** As mentioned in Section 3, we proposed the residual adaptor (RA), which can extract the task-specific compensation feature from the low-level features, to deal with the discrepancy issue.

From the result shown in Fig. 2, adding the task-specific RA in the KDMTS model obtains better performance on indoor localization and user identification than that it. However, it does not boost the performance of gesture recognition. As for this, we think the reason is that there is a lack of supervised information to train the RA. Since the loss of KDMTS is only related to the common feature extraction and the final predicted labels, we introduce the logits distillation to supply enough information for training.

**Analysis of logits distillation.** As mentioned above, we considered that the supervised information is not enough to train an excellent RA. Then we add logits distillation to sufficiently exploit the trained STS models and enrich the supervised information for Wimuse, which is proved to be extremely efficient.

As presented in Fig. 2, it is clear that Wimuse, adding the logits distillation, achieves a significantly better overall performance. As we expected, the logits distillation brought more information from the trained STS models to Wimuse for better performance.

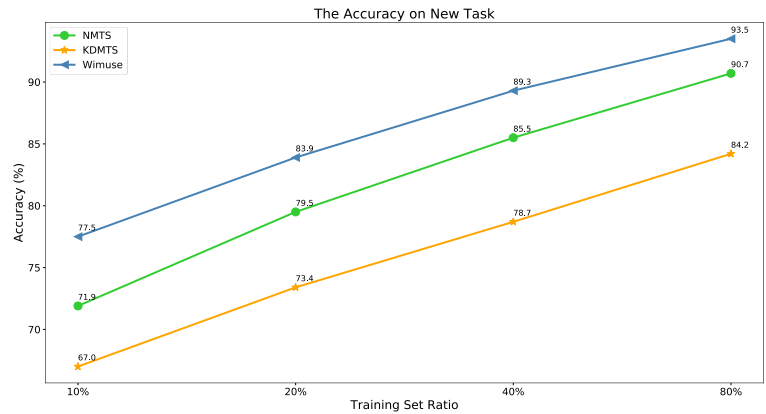

**Fig. 3.** The effects of the task-specific residual adaptor and the logits distillation

**Analysis of task scalability.** We conduct experiments on CSIDA and Widar3.0 datasets to evaluate the task scalability of Wimuse. Specifically, we construct the three-task version model by adding some modules to the pre-trained two-task version model. Then, we trained the added parts of the three-task version with the new task.

As shown in figure 3, the accuracies of NMTS, KDMTS, and Wimuse increase steadily along with the growth of the training set ratio. Further, Wimuse achieves higher accuracy than the other two models, which shows that Wimuse has better task scalability than other models. And the learned common features are more general than those in other models.

Table 5. The comparison of different methods.

| Type | Method | No. Parameters | No. Mult-adds (Million) | Memory (MB) | Time (ms) |
|---|---|---|---|---|---|
| STS | For GR task | 674,566 | 285.56 | 16.18 | 46.243 |
|  | For IL task | 677,136 | 285.60 | 16.19 | 45.776 |
| MTS | ARIL | 3,490,246 | 480.82 | 29.02 | 50.714 |
|  | NMTS | 563,222 | 298.20 | 16.00 | 46.767 |
|  | UMTS | 563,222 | 298.20 | 16.00 | 46.811 |
|  | KDMTS | 563,222 | 298.20 | 16.00 | 45.831 |
|  | KDMTS (with RA) | 662,038 | 411.45 | 21.12 | 46.243 |
|  | Wimuse (ours) | 662,038 | 411.45 | 21.12 | 46.243 |

'No. parameters' denotes the number of the trainable parameters. 'No. Multi-adds' represents the number of multiplexes and addition operations in labeling a new sample. 'Memory' and 'Time' are the storage memory and time needed in labeling a new sample, respectively.

**Analysis of model complexity and computational cost.** We also compare the model complexity, memory consumption, and runtime under the joint task of gesture recognition and indoor localization on the ARIL dataset. The input size is 52×192 with batch size as 16. As table 5 demonstrated, the MTS methods (except the ARIL method) have fewer parameters, operations, memory consumption, and runtime. It proves that MTS methods consume less, but is more efficient. In addition, compare with other MTS methods, KDMTS (with RA) and Wimuse increased the computational cost and model complexity. But it is still better than ARIL on computational cost and model complexity.

## 5    Conclusion

In this paper, we propose a WiFi-based multi-task sensing model (Wimuse) to perform gesture recognition, user identification, and indoor localization simultaneously. First, We reveal the imbalance and discrepancy issues in WiFi-based MTS. Then, we adopt the knowledge distillation technique and task-specific residual adaptor to address these issues. Next, we conduct comprehensive experiments on three public datasets (i.e., the ARIL dataset, the CSIDA dataset, and the Widar3.0 dataset). The evaluation suggests that Wimuse achieves state-of-the-art performance with the average accuracy of 85.20%, 98.39%, and 98.725% on the joint task of gesture recognition, indoor localization, and user identification task respectively.

Though we get satisfying results on Wimuse, there exists still two aspects for improvement, which are that we make no use of the phase data of CSI and we just achieve three tasks simultaneously. There are several fruitful directions for future investigation. i) Wimuse adopt CNN to extract features from the CSI samples. The CSI is time-series data while CNN is not suitable for sequence data. We need to adjust Wimuse to adapt the streaming data. ii) There are other more tasks that we need to exploit for the WiFi-based MTS, such as breath detection, user orientation estimation, fall detection. iii) Since the CSI can be captured from a commercial WiFi device, we can try to deploy Wimuse into the WiFi device. And this requires a lightweight design of Wimuse to satisfy the limitations of memory, computing power in WiFi devices. iv)We can try to utilize the phase data of CSI by developing a general phase denoise method to achieve better performance.